\documentclass[11pt]{article}

\usepackage[final]{acl}

\usepackage{times}
\usepackage{latexsym}

\usepackage[T1]{fontenc}

\usepackage[utf8]{inputenc}

\usepackage{microtype}

\usepackage{inconsolata}

\usepackage{graphicx}

\usepackage{tikz}
\usetikzlibrary{arrows.meta, positioning, shapes.geometric}
\usepackage{amsmath} 
\usepackage{url}
\usepackage{pgfplots}
\pgfplotsset{compat=1.17}
\usetikzlibrary{positioning,arrows.meta}
\usepackage{booktabs}
\usetikzlibrary{calc}

\title{SalamandraTA at WMT 2026 Terminology Shared Task: Hard Examples Are Better Teachers}

\author{Xixian Liao \\
  Barcelona Supercomputer Center \\
  \texttt{xixian.liao@bsc.es} \\\And
  Maite Melero \\
  Barcelona Supercomputer Center \\
  \texttt{maite.melero@bsc.es} \\}

\begin{document}
\maketitle
\begin{abstract}

Terminology-aware translation asks for more than a correct translation: the output must use the exact terms a glossary prescribes. The standard recipe, fine-tuning on glossary-annotated translation pairs, hides an inefficiency: for most examples the glossary prescribes exactly what the model would have produced anyway, so they teach nothing about following a glossary. We therefore keep only the examples where the model's own translation contradicts the glossary.
In a controlled study at fixed data volume, this selection alone raises term accuracy from 78.7\% to 89.9\%.
The filtered data, built by a two-way synthetic pipeline on open models, is part of the instruction-tuning mixture of our public release \textsc{SalamandraTA-7b-instruct} v3.0, which, used exactly as released
and wrapped in a document-level inference pipeline, forms the BSC submission to the WMT26 Terminology Shared Task Track~1.
At the official WMT26 evaluation, our system achieves 94.2\% term success at 74.6 chrF++, with only two of the twenty-two submissions outperforming it on both metrics. On last year's benchmark, it also surpasses our GRPO-based system, despite being trained solely with ordinary supervised fine-tuning.

\end{abstract}

\section{Introduction}

\begin{figure}[t]
  \centering
  \includegraphics[width=0.47\textwidth]{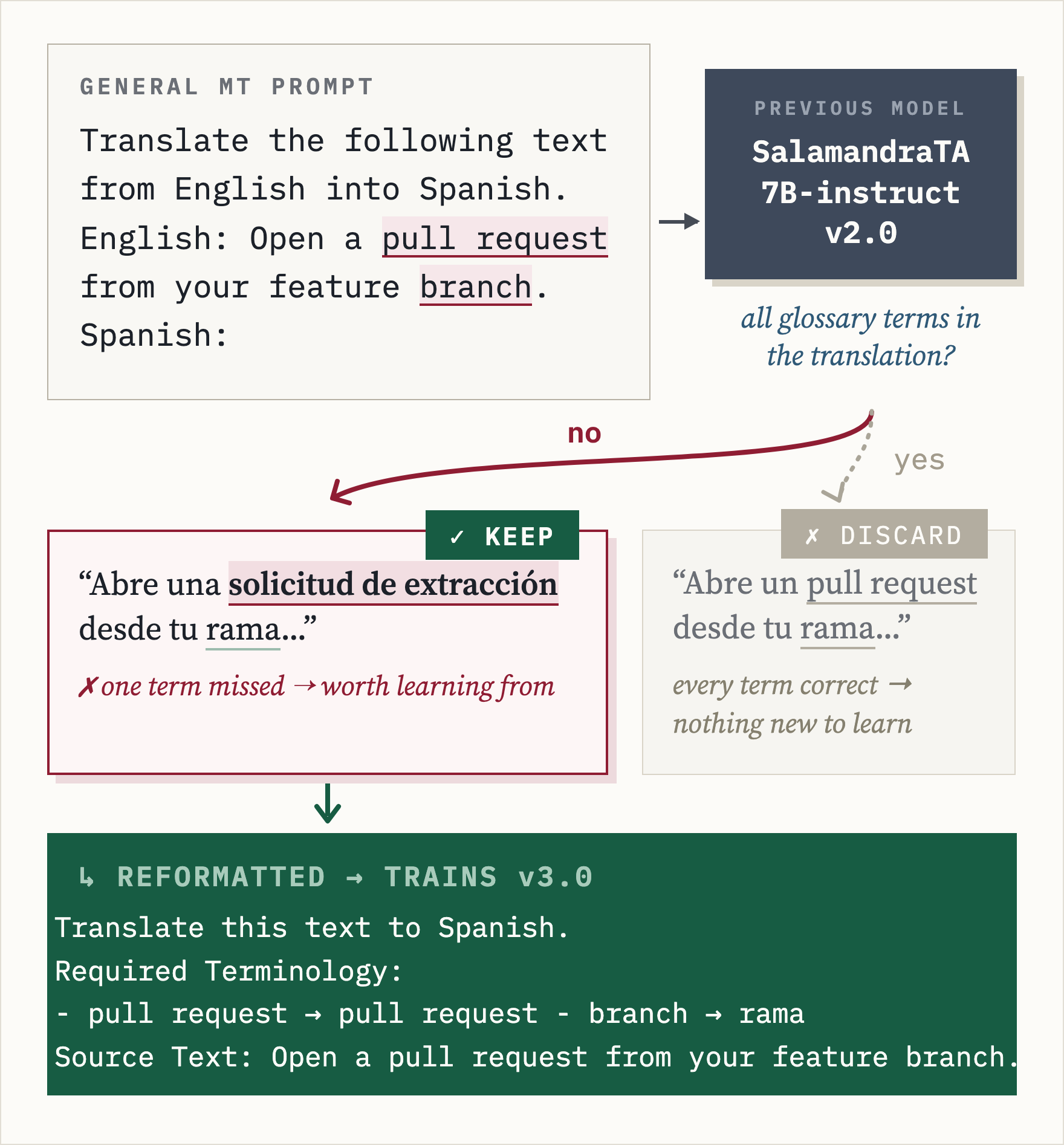}
  \caption{Data selection: train only on what \textsc{SalamandraTA-7B-instruct} gets wrong.}
  \label{fig:method}
\end{figure}

A translation can preserve the meaning perfectly and still fail its audience. A Spanish-speaking developer who opens a \emph{pull request} will call it exactly that, yet a generic machine translation (MT) system may render the phrase as \emph{solicitud de extracción}.
Both are understandable, but only the English term matches what developers read every day in their interfaces and documentation.
Terminology-aware translation addresses this gap: beyond producing a correct translation, it must produce the prescribed one, whether the prescription comes from a client's glossary or from the usage of a community.

A common way to teach this behaviour is to fine-tune the model on glossary-annotated translation pairs.
Much of this supervision, however, is redundant: for a large share of training examples the glossary prescribes exactly what the model would have produced anyway, so the example can be answered correctly without consulting the glossary at all.
The useful signal is concentrated in the cases where the prescribed term differs from the model's default choice.
This observation suggests a simple data selection rule: keep for training only those where at least one term is hard for the model and comes out wrong (Figure~\ref{fig:method}).
Our controlled experiment confirms that the rule matters.
Holding the training budget and every other ingredient fixed, we vary only the proportion of these hard examples in the terminology data, and term accuracy climbs from 78.7\% with none of them to 89.9\% with all of them (Figure~\ref{fig:ratio-curve}).

This paper describes the Barcelona Supercomputing Center (BSC) submission to the WMT26 Terminology Translation Shared Task Track~1, document-level translation with an explicit dictionary. 
Our system is our latest public release, \textsc{SalamandraTA-7B-instruct} v3.0,\footnote{\url{https://huggingface.co/BSC-LT/salamandraTA-7b-instruct}} used exactly as released: its instruction-tuning mixture already includes terminology data built with this recipe, generated by a two-way synthetic pipeline on Gemma-4-31B \citep{gemmateam2026gemma4} and filtered against our previous release.
At inference time, documents are translated chunk by chunk with per-chunk glossaries and lightly post-edited under a guard that protects prescribed terms.
At the official WMT26 evaluation, our system achieves 94.2\% term success at 74.6 chrF++, with only two of the twenty-two submissions outperforming it on both metrics. On last year's benchmark, it also surpasses our GRPO-based system, despite being trained solely with ordinary supervised fine-tuning.

In summary, we make three contributions.
First, we show that a lot of glossary-annotated training data is redundant, and that a simple remedy, keeping only the examples where the glossary contradicts the model's default translation, yields an eleven-point term-accuracy gain at fixed data volume.
Second, we describe a two-way synthetic pipeline built on open models, which turns glossaries into documents and documents into glossaries, and whose output trains the terminology component of the public release \textsc{SalamandraTA-7b-instruct} v3.0.
Third, we present the BSC Track~1 submission: the released model, unmodified, wrapped in a document-level inference pipeline with per-chunk glossaries and a post-editing guard that protects prescribed terms.

\section{Related work}

\paragraph{Making models follow glossaries.}
The terminology constraint was earlier imposed at decoding time, through lexically constrained beam search \citep{hokamp-liu-2017-lexically}, at a cost in speed and fluency.
\citet{dinu-etal-2019-training} moved it into training, annotating source terms with their prescribed translations so the model learns to copy them in context.
The WMT terminology shared tasks \citep{alam-etal-2021-findings, semenov-etal-2023-findings, semenov-etal-2025-findings} record the field's shift to LLMs.
Our own previous submission moved the constraint into the reward, optimising models with GRPO under a joint adherence and quality objective \citep{garcia-gilabert-etal-2025-terminology}.
This year the recipe is simpler: ordinary supervised fine-tuning, targeting longer document inputs. What is new is not how we train but what we train on: only the examples whose terminology the model gets wrong.

\paragraph{Synthetic terminology data.}
LLM-generated parallel text now rivals web-crawled corpora in quality \citep{finkelstein-etal-2024-introducing}, and strong WMT25 terminology systems already relied on curated synthetic data and LLM post-editing \citep{jaswal-2025-takes, semenov-etal-2025-findings}.
Our pipeline differs in shape and in openness: it runs in both directions, expanding glossaries into documents and mining documents for glossaries, and is built entirely on openly licensed models and data.

\paragraph{Choosing what to train on.}
That some examples teach more than others is a recurring observation. 
In MT, selection began as domain adaptation \citep{axelrod-etal-2011-domain}, later made dynamic during training
\citep{van-der-wees-etal-2017-dynamic}; curriculum learning orders examples by difficulty instead of filtering them \citep{bengio-2009, kocmi-bojar-2017-curriculum, platanios-etal-2019-competence}. The instruction-tuning era sharpened the point that a small, well-chosen set can beat a large undifferentiated one \citep{zhou2023limaalignment, chen2024alpagasustrainingbetteralpaca, liu2024makesgooddataalignment}, and, closest to us, \citet{kocmi-etal-2025-command} filter MT data by difficulty before preference tuning.
Concentrating supervision where the current model falls short is also the animating idea of active learning \citep{settles-2009-survey}, long used in MT to decide what to annotate \citep{haffari-etal-2009-active, zhao-etal-2020-active}.
In our case, we select from an already-labelled pool, so no annotator is queried; and where prior selection relies on a continuous score, however obtained, our notion of hardness is self-referential and binary: an example is hard exactly when the model's own unconstrained output contradicts the prescribed term.
The filter itself has no scores to threshold, and \S\ref{sec:ablation} measures its individual contribution.

\section{Data: two synthetic pipelines}
\label{sec:data}

The terminology component of \textsc{SalamandraTA} v3's instruction-tuning mixture is built by two synthetic pipelines that run in opposite directions. One starts from terminology and generates text: bilingual medical glossaries \citep[MeSpEn;][]{villegas2018mespen} seed Gemma-4-31B\footnote{\url{https://huggingface.co/google/gemma-4-31B}} \citep{gemmateam2026gemma4}, which writes parallel texts around them.
The other starts from text and extracts terminology: Gemma-4-31B extracted aligned term pairs from the EMEA parallel corpus\footnote{\url{https://opus.nlpl.eu/datasets/EMEA}} \citep{tiedemann-2012-parallel}, whose sentences we then assemble
into documents.
Both pipelines share one selection step, described in \S\ref{sec:filter}, that keeps only what our own model gets wrong.
Together they yield 33{,}615 instances covering 94 directed pairs across 29 languages (12{,}479 from MeSpEn, 21{,}136 from EMEA); Appendix~\ref{app:synthetic-pipeline} documents cleaning rules, generation prompts, and filtering thresholds (Figure~\ref{fig:pipeline} gives an overview).

\subsection{MeSpEn: terminology \textrightarrow{} text}
\label{sec:mespen}

After cleaning the raw glossaries, we keep only the hard terms, those our model mistranslates in isolation (\S\ref{sec:filter}), and use them as seeds.
Gemma-4-31B writes a parallel text around small groups of seeds, grouped by semantic similarity so that co-occurring terms
are plausible together, at lengths ranging from a single sentence to multi-paragraph documents.
In document-length samples the main term must appear at least twice and be translated identically each time, mirroring the consistency requirement of the shared task.
A sample survives only if every annotated term actually appears on both sides, and term and sentence pairs exceed LaBSE \citep{feng-etal-2022-language} similarity thresholds.

\subsection{EMEA: text \textrightarrow{} terminology \textrightarrow{}
documents}
\label{sec:emea}

Gemma-4-31B extracts aligned term pairs from each EMEA sentence pair, and sentences on which our model already produces every gold term are discarded (\S\ref{sec:filter}). After quality filtering, deduplication, and LaBSE filtering, consecutive sentences are concatenated into a mixture of granularities: single sentences, paragraphs, and multi-paragraph documents with literal newline breaks, each document's glossary being the union of its sentences' term pairs.
Finally, glossary entries are reduced to lemma form while the text is left untouched, so the model must learn to inflect prescribed
terms in context rather than copy surface strings.

\subsection{The hard-example filter}
\label{sec:filter}

Both pipelines rely on the same selection step, and it is the step this paper is about.
Every candidate, a glossary term in MeSpEn or a sentence in EMEA, is translated by \textsc{SalamandraTA-7b-instruct} v2.0 \emph{without} any terminology in the prompt, and the output is checked against the gold terms with a lenient matcher that forgives casing, punctuation, and spacing differences.
If every required term already appears, the model needs no help on this example and it is discarded; only candidates with at least one missed term survive. We call the discarded examples \emph{easy} and the surviving ones \emph{hard}.
The filter is computed once, against our previous public release.


\section{Controlled study: does hardness matter?}
\label{sec:ablation}

\begin{figure}[t]
  \centering
  \includegraphics[width=0.49\textwidth]{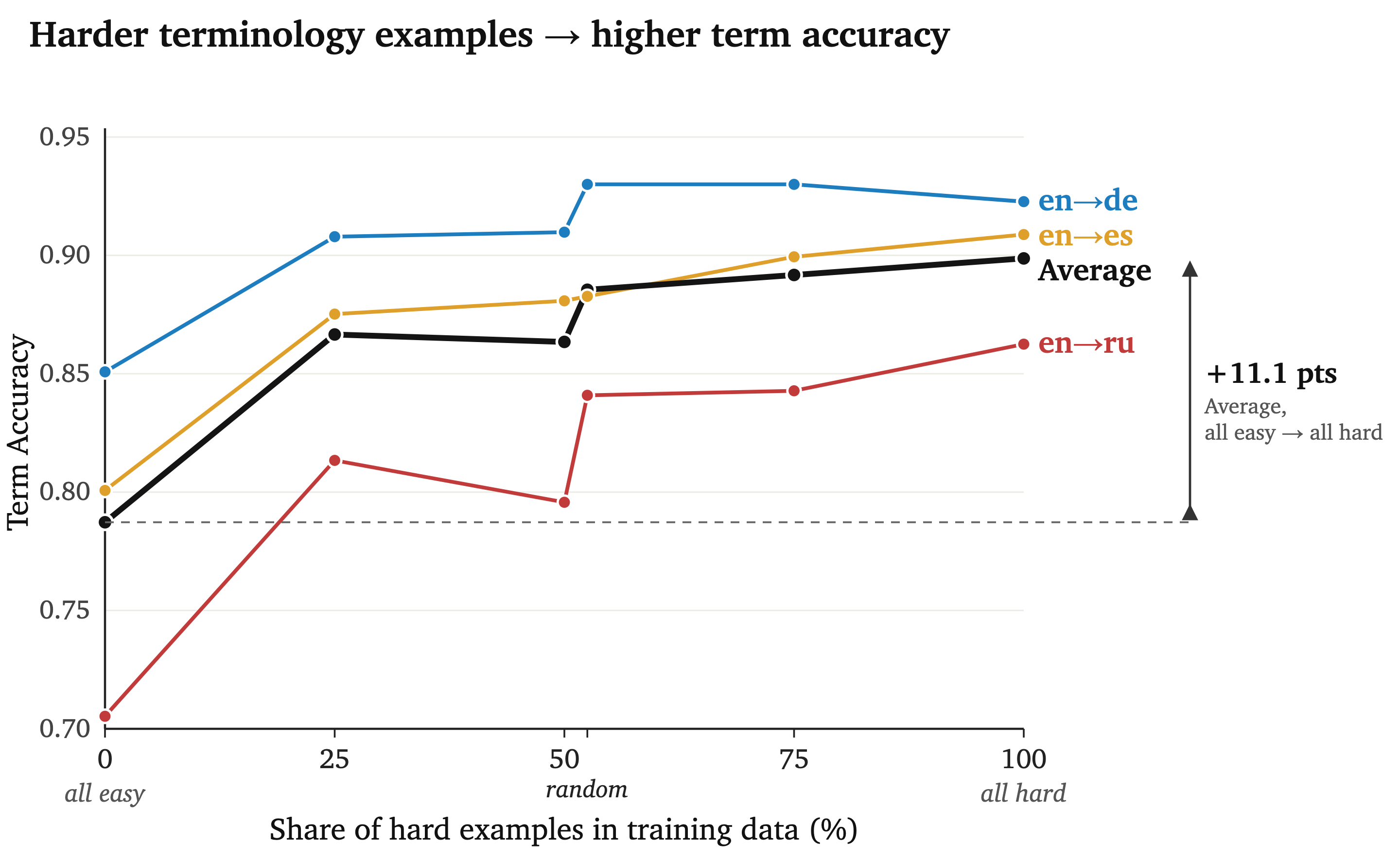}
  \caption{Term accuracy against the share of hard examples in the terminology data.}
  \label{fig:ratio-curve}
\end{figure}

The filter of \S\ref{sec:filter} discards nearly half of the corpus (47.5\%; Appendix~\ref{app:pools}).
This section asks whether that is actually a good idea. We hold everything about training fixed and vary a single quantity, the share of hard examples in the terminology data, from 0\% to 100\% in five steps plus a random baseline, and trace term accuracy along the way (Figure~\ref{fig:ratio-curve}).

\subsection{Setup}
\label{sec:ablation-setup}

\paragraph{Training runs.}
In Figure~\ref{fig:ratio-curve}, each point on the curve is a complete instruction-tuning run.
Starting from the \textsc{SalamandraTA-7B} base model (the continually pre-trained translation checkpoint), we fine-tune on the fixed-background mixture of the public \textsc{SalamandraTA-7b-instruct} v2 release recipe (see data datails in Appendix~\ref{app:it-mixture}).
All runs use identical hyperparameters: one epoch, learning rate $10^{-5}$ with cosine decay and 3\% warmup, context length 8{,}192, and an effective batch size of 32.

\paragraph{The six mixtures.}
Into this background we insert a terminology component, the only ingredient that changes between runs.
It spans the 19 directions the EMEA branch produces, with a fixed per-direction budget of up to 4{,}000 examples, identical across mixtures.
Six compositions are tested: 0\%, 25\%, 50\%, 75\%, and 100\% hard examples, plus a \emph{random} baseline sampled from the unfiltered pool, which inherits the corpus's natural per-direction proportion of hard examples.
Easy examples and hard examples pass the same cleaning, quality, and semantic-similarity gates, so the two pools differ only in difficulty.

\paragraph{Evaluation.}
We evaluate on the WMT2025 Terminology Track~1 test set \citep{semenov-etal-2025-findings}: en$\rightarrow$de, en$\rightarrow$es, and en$\rightarrow$ru.\footnote{\url{https://github.com/wmt-conference/wmt25-terminology}}
None of the three occurs as a training direction in the terminology component, so any gain on them already reflects transfer.
We report the official term accuracy, micro-averaged over all 1{,}589 term occurrences, together with chrF \citep{popovic-2015-chrf} as a check that translation quality does not silently degrade; confidence intervals and $p$-values come from a clustered paired bootstrap over source segments (95\%, 10k resamples).

\paragraph{Relation to the released model.}
\label{paragraph:diff_v2_v3}
These runs do not reproduce the v3.0 recipe. The background mixture follows the earlier v2.0 release recipe and does not include the additional tasks introduced in v3.0 (structured-text translation, register control, paraphrasing, and further post-editing and gender data). The terminology component also differs from the released one: v3.0 is trained on MeSpEn and EMEA data together, whereas each run here uses EMEA data only, at a fixed volume of 76{,}000 instances, so that the hard-example ratio is the only moving part across runs. The curve should accordingly be read as evidence for the selection principle, not as a decomposition of v3.0's gains.

\subsection{Results}
\label{sec:ablation-results}

Selection alone is worth eleven points.
Figure~\ref{fig:ratio-curve} shows the result, with per-direction numbers in Appendix~\ref{app:ablation-numbers}: averaged over the three directions, term accuracy climbs from 78.7\% with no hard examples to 89.9\% with only hard examples, a gain of 11.2 points (95\% CI $[+9.5, +12.8]$, $p<0.0001$) at identical data volume.
The easy examples are not noise; they passed exactly the same cleaning, quality, and similarity gates as the hard ones.

Accuracy generally follows the dose of difficulty across every mixture we tested. 
Hard examples are potent: a quarter dose already delivers 71\% of the full effect. 
The small dip at 50\%, traced to the en$\rightarrow$ru direction, lies within noise on the average (25\% vs.\ 50\%: $+0.3$ points, 95\% CI $[-0.6, +1.3]$, $p=0.55$, paired bootstrap).
A forced 50/50 split trails the fully filtered set by 3.5 points.
An unfiltered random sample of the same size comes closer, but for a revealing reason: EMEA is naturally rich in hard sentences, so a random draw is itself a moderately hard mixture, and it lands where its hard share predicts.
Even against this lucky corpus, filtering adds a significant $+1.3$ points ($p=0.003$); on a corpus with fewer hard cases, random sampling would inherit correspondingly less, while the filter turns the dose of difficulty into a guarantee.

The strongest evidence that difficulty is what does the teaching comes from the direction the training data covers worst. For en$\rightarrow$ru, with no Russian anywhere in the terminology data, the easy-to-hard gain is the largest of the three ($+15.7$ points).
The en$\rightarrow$es curve climbs strictly with the hard share ($+10.8$), and en$\rightarrow$de, the direction closest to the training languages, saturates earliest, with 75\% hard and the random mixture edging past pure-hard (93.0 vs.\ 92.3).

Hardness costs nothing in quality.
Across the five mixtures that contain hard examples, average chrF stays within half a point (67.0--67.4), and the all-easy mixture is the weakest on chrF as well (66.5). The gains are not bought with translation quality.

\section{Document-level submission pipeline}
\label{sec:inference}

The WMT26 Track~1 test sets are full documents, each paired with a corpus-level glossary that mixes entries relevant to the document with distractors from the whole collection.
The 8{,}192-token context window of our system must hold the document, its glossary, and the translation as it is generated, so full documents invite truncation and repetition; and oversized glossaries dilute attention over entries that never occur in the text. 
The submission therefore translates each document in three stages.

\paragraph{1. Layout-preserving chunking.}
A document is split only at line breaks into chunks of roughly 500 source words, translated chunk by chunk with the terminology prompt of the model card (beam search, beam size 5), and re-joined into documents.
A chunk spans several lines, and the model reproduces its line structure directly, having been trained on exactly these granularities (\S\ref{sec:data}). 
An automatic check verifies every chunk, and the few that call for it are re-translated line by line, so the submitted layout matches the source by construction.

\paragraph{2. Glossary filtering.}
Each chunk receives only the glossary entries whose source term actually occurs in it.
A term counts as present if it appears verbatim or if its lemmas match.
This shrinks the corpus-level glossary to the handful of entries the model must act on.

\paragraph{3. QE-guided post-editing.}
Chunk translation leaves occasional local defects: dropped clauses, agreement errors, spelling noise.
We therefore score every segment with CometKiwi \citep{rei-etal-2022-cometkiwi} and let the same released model post-edit the low-scoring ones, using the post-editing prompt from its instruction tuning, so that the submission comes entirely from one model.

The risk is that fluency and terminology disagree about repetition.
The test documents often name one concept several ways while the glossary maps every name to the same target, so faithful translations repeat themselves: \emph{a steel wheel (or steelie)} becomes \emph{ko\l{}o stalowe (lub ko\l{}o stalowe)}.
The post-editor does what any writer would and smooths the repetition away.
Sometimes this is harmless, because the document keeps enough copies of the term; sometimes it deletes an occurrence that term accuracy still needed, or replaces the prescribed wording altogether.
CometKiwi rewards all three alike.
An edit is therefore applied only if it raises CometKiwi and does not lower document-level term accuracy.
Most edits that survive fix grammar, spelling, or punctuation, or remove a genuinely surplus repetition.

\section{Results}
\label{sec:results}

\subsection{Official WMT26 results}

Table~\ref{tab:wmt26} places \textsc{SalamandraTA-7b-instruct} v3.0, used exactly as released and wrapped in the pipeline of \S\ref{sec:inference}, among the WMT26 Track 1 submissions \citep{charkiewicz-etal-2026-findings}.
The system reaches 94.2\% lemmatised term success at 74.6 document chrF++, only 2.2 points of term success below the human reference.
Of the twenty-two submissions, only \textsc{Cozy} and Agentic-OPUS improve on both axes at once; four further systems reach higher term success at lower chrF++.

\begin{table*}[t]
\centering\small
\setlength{\tabcolsep}{4pt}
\begin{tabular}{lcccccc}
\toprule
& \multicolumn{3}{c}{Term success} & \multicolumn{3}{c}{chrF++} \\
\cmidrule(lr){2-4}\cmidrule(lr){5-7}
System & Avg & en-pl & es-eu & Avg & en-pl & es-eu \\
\midrule
\textit{gold} & \textit{96.4} & \textit{93.6} & \textit{99.1} & \textit{100.0} & \textit{100.0} & \textit{100.0} \\
\midrule
Agentic-OPUS & \textbf{95.7} & 95.4 & 96.0 & 75.3 & \textbf{77.1} & 73.5 \\
\textsc{Cozy}$_{\text{flash}}$ & 95.4 & 94.4 & 96.3 & 74.4 & 76.5 & 72.2 \\
HW-TSC & 95.3 & 95.8 & 94.8 & 73.8 & 75.6 & 72.1 \\
\textsc{Cozy} & 95.0 & 93.6 & \textbf{96.4} & \textbf{75.6} & 75.9 & \textbf{75.3} \\
Agentic-Sonnet & 94.7 & 95.2 & 94.3 & 73.1 & 75.6 & 70.7 \\
TaT & 94.5 & \textbf{97.2} & 91.7 & 73.7 & 73.6 & 73.8 \\
\textbf{SalamandraTA v3.0 (ours)} & 94.2 & 93.2 & 95.2 & 74.6 & 75.9 & 73.3 \\
CUNI-UFAL & 93.4 & 91.6 & 95.2 & 74.4 & 74.2 & 74.5 \\
\bottomrule
\end{tabular}
\caption{WMT26 Track 1, proper mode: lemmatised exclusive term success (\%) and document chrF++, per direction and averaged. The eight submissions with the highest average term success, out of twenty-two; best among these in bold. Full table in the findings paper \citep{charkiewicz-etal-2026-findings}.}
\label{tab:wmt26}
\end{table*}

\subsection{Comparison with our GRPO submission}

Our WMT25 system optimised the same model family with GRPO under a joint adherence and quality reward \citep{garcia-gilabert-etal-2025-terminology}. It operates only at the sentence level, so the two can only be compared on last year's benchmark (Table~\ref{tab:wmt25}).
The released v3.0 model improves on both axes, from 67.3 to 69.4 average chrF and from 91.3\% to 94.0\% average term accuracy, with no reinforcement learning, no task-specific adaptation, and no access to the test-time domain.
The comparison is not fully controlled, however. The two systems differ not only in their terminology data, but also in the broader instruction mixture. We therefore view this as a system-level comparison rather than a controlled ablation.
Nevertheless, the result does show that a general-purpose translation model, given our terminology data as part of ordinary instruction tuning, can outperform a dedicated RL pipeline on this task.

\begin{table}[t]
\centering\small
\begin{tabular}{lcc}
\toprule
System (WMT25 Track 1) & chrF & Term \\
\midrule
o3-term-guide & 71.0 & 99.1 \\
duterm & 70.1 & 98.2 \\
\textbf{SalamandraTA v3.0 (ours)} & 69.4 & 94.0 \\
salamandrata (ours, GRPO) & 67.3 & 91.3 \\
\bottomrule
\end{tabular}
\caption{WMT25 Track 1, averaged over en$\rightarrow$es/de/ru: our released model against our previous GRPO submission and the two top-ranked systems \citep{semenov-etal-2025-findings}.}
\label{tab:wmt25}
\end{table}

\section{Discussion}
\label{sec:discussion}

\paragraph{Many-to-one glossaries make quality and adherence disagree.}
The test glossaries are not injective: \emph{boot lid} and \emph{trunk lid} share the single prescribed target \emph{klapa baga\.znika}.
When a source document deliberately enumerates synonyms, a faithful translation must repeat that target: ``both the boot lid and the trunk lid'' can only become \emph{zar\'owno klapa baga\.znika, jak i klapa baga\.znika}, text any human editor would immediately smooth out.
Quality estimation sides with the editor: CometKiwi scored the smoothed variant, \emph{... jak i pokrywa baga\.znika}, 0.13 higher, and across our post-editing stage it consistently assigned positive deltas to the very edits that damage adherence.
We suspect this tension generalises: any system that optimises a quality metric on top of a terminology task needs an explicit adherence constraint, because the two objectives are locally adversarial on some many-to-one entries.
What the translations of these sentences should ideally look like, and how they should be scored in the terminology task, remain open questions.

\paragraph{What the filter buys.}
The controlled study suggests the value of hard examples is not the term pairs they contain but the behaviour they force: reading the glossary when the model's default would have been something else.
This also suggests the filter should transfer to other adherence-flavoured tasks (style guides, register constraints, do-not-translate lists) where most naturally occurring supervision is similarly redundant.
Whether this behavioural account holds up mechanistically, and what in the model implements it, we leave to future work.

\section{Conclusion}
We described the BSC submission to WMT26 Terminology Translation Track~1: the public release \textsc{SalamandraTA-7b-instruct} v3.0, whose terminology training data is built by a two-way synthetic pipeline on open models and filtered by a single rule, keep only what the model gets wrong, together with a document-level inference pipeline with per-chunk glossaries and QE-guided post-editing.
A controlled study shows the filter is responsible for an eleven-point term-accuracy gain at fixed data volume, and the released system reaches 94.2\% term success at 74.6 chrF++ in the WMT26 official evaluation.
Hard examples, it turns out, are better teachers.

\section*{Limitations}

The WMT25 test set provides per-sentence terminology, so its errors are all genuine adherence failures.
The WMT26 setting differs: a corpus-level glossary must first be narrowed to each chunk by surface and lemma matching (\S\ref{sec:inference}), and a term whose inflected form escapes the match never reaches the model at all. Our submission's errors will therefore mix adherence failures with retrieval failures.
The controlled study is deliberately narrow: one data source, three test directions, one data volume, and a single reference model defining hardness.
The filter is also static, computed once against v2.0 before training; a dynamic variant that re-estimates difficulty during training might select better but was out of scope.
Our terminology data is biased toward the medical domain of its seed resources, while the test domains this year differ. 
Finally, the base model was continually pre-trained on sentence-level parallel data only. Document-level translation behaviour comes primarily from instruction tuning, and very long documents must be chunked at inference, which can lose cross-chunk context such as antecedents for consistent term choice.
We are currently extending continual pre-training to document-level parallel corpora, which we expect will improve document-level context modeling.

\section*{Acknowledgments}
This work/research has been promoted and financed by the Government of Catalonia through the Aina project.

This work is supported by MLLM4TRA (PID2024-158157OB-C32) funded by MCIN/AEI/10.13039/501100011033/FEDER, UE.

This work was supported by the TaMTAS project (PCI2025-167117-2), funded by MICIU/AEI/10.13039/501100011033 and the European Union under the CHIST-ERA Call 2025, \textit{Science in Your Own Language}.

\bibliography{custom}

@inproceedings{charkiewicz-etal-2026-findings,
    title = "Findings of the {WMT}26 Terminology Translation Task",
    author = "Charkiewicz, Adrian  and
      Chen, Pinzhen  and
      Etchegoyhen, Thierry  and
      Gete Ugarte, Harritxu  and
      Guttmann, Kamil  and
      Huang, Xu  and
      Ponce, David  and
      Nowakowski, Artur  and
      Odermatt, Fr{\'e}d{\'e}ric  and
      Oncevay, Arturo  and
      Semenov, Kirill  and
      Zhu, Dawei  and
      Zouhar, Vil{\'e}m",
    booktitle = "To appear in Proceedings of the Eleventh Conference on Machine Translation",
    year = "2026",
    publisher = "Association for Computational Linguistics",
}

@inproceedings{zhao-etal-2020-active,
    title = "Active Learning Approaches to Enhancing Neural Machine Translation",
    author = "Zhao, Yuekai  and
      Zhang, Haoran  and
      Zhou, Shuchang  and
      Zhang, Zhihua",
    editor = "Cohn, Trevor  and
      He, Yulan  and
      Liu, Yang",
    booktitle = "Findings of the Association for Computational Linguistics: EMNLP 2020",
    month = nov,
    year = "2020",
    address = "Online",
    publisher = "Association for Computational Linguistics",
    url = "https://aclanthology.org/2020.findings-emnlp.162/",
    doi = "10.18653/v1/2020.findings-emnlp.162",
    pages = "1796--1806"
}

@inproceedings{haffari-etal-2009-active,
    title = "Active Learning for Statistical Phrase-based Machine Translation",
    author = "Haffari, Gholamreza  and
      Roy, Maxim  and
      Sarkar, Anoop",
    editor = "Ostendorf, Mari  and
      Collins, Michael  and
      Narayanan, Shri  and
      Oard, Douglas W.  and
      Vanderwende, Lucy",
    booktitle = "Proceedings of Human Language Technologies: The 2009 Annual Conference of the North {A}merican Chapter of the Association for Computational Linguistics",
    month = jun,
    year = "2009",
    address = "Boulder, Colorado",
    publisher = "Association for Computational Linguistics",
    url = "https://aclanthology.org/N09-1047/",
    pages = "415--423"
}

@techreport{settles-2009-survey,
  title={Active Learning Literature Survey},
  author={Settles, Burr},
  institution={University of Wisconsin-Madison Department of Computer Sciences},
  year={2009}
}

@article{mashParaCLEANImprovingTranslation,
  title = {{{ParaCLEAN}}: {{Improving Translation Quality}} through {{Systematic Parallel Data Cleaning}}},
  author = {Mash, Audrey and Bohman, Ella Paulina and Melero, Maite},
  year = 2026,
  journal = {Proceedings of the Fifteenth Language Resources and Evaluation Conference (LREC 2026)},
  pages = {6630--6640},
  publisher = {ELRA Language Resources Association (ELRA),},
  langid = {english}
}

@inproceedings{rei-etal-2022-cometkiwi,
    title = "{C}omet{K}iwi: {IST}-Unbabel 2022 Submission for the Quality Estimation Shared Task",
    author = "Rei, Ricardo  and
      Treviso, Marcos  and
      Guerreiro, Nuno M.  and
      Zerva, Chrysoula  and
      Farinha, Ana C  and
      Maroti, Christine  and
      C. de Souza, Jos{\'e} G.  and
      Glushkova, Taisiya  and
      Alves, Duarte  and
      Coheur, Luisa  and
      Lavie, Alon  and
      Martins, Andr{\'e} F. T.",
    editor = {Koehn, Philipp  and
      Barrault, Lo{\"i}c  and
      Bojar, Ond{\v{r}}ej  and
      Bougares, Fethi  and
      Chatterjee, Rajen  and
      Costa-juss{\`a}, Marta R.  and
      Federmann, Christian  and
      Fishel, Mark  and
      Fraser, Alexander  and
      Freitag, Markus  and
      Graham, Yvette  and
      Grundkiewicz, Roman  and
      Guzman, Paco  and
      Haddow, Barry  and
      Huck, Matthias  and
      Jimeno Yepes, Antonio  and
      Kocmi, Tom  and
      Martins, Andr{\'e}  and
      Morishita, Makoto  and
      Monz, Christof  and
      Nagata, Masaaki  and
      Nakazawa, Toshiaki  and
      Negri, Matteo  and
      N{\'e}v{\'e}ol, Aur{\'e}lie  and
      Neves, Mariana  and
      Popel, Martin  and
      Turchi, Marco  and
      Zampieri, Marcos},
    booktitle = "Proceedings of the Seventh Conference on Machine Translation (WMT)",
    month = dec,
    year = "2022",
    address = "Abu Dhabi, United Arab Emirates (Hybrid)",
    publisher = "Association for Computational Linguistics",
    url = "https://aclanthology.org/2022.wmt-1.60/",
    doi = "10.18653/v1/2022.wmt-1.60",
    pages = "634--645"
}

@inproceedings{popovic-2015-chrf,
    title = "chr{F}: character n-gram {F}-score for automatic {MT} evaluation",
    author = "Popovi{\'c}, Maja",
    editor = "Bojar, Ond{\v{r}}ej  and
      Chatterjee, Rajan  and
      Federmann, Christian  and
      Haddow, Barry  and
      Hokamp, Chris  and
      Huck, Matthias  and
      Logacheva, Varvara  and
      Pecina, Pavel",
    booktitle = "Proceedings of the Tenth Workshop on Statistical Machine Translation",
    month = sep,
    year = "2015",
    address = "Lisbon, Portugal",
    publisher = "Association for Computational Linguistics",
    url = "https://aclanthology.org/W15-3049/",
    doi = "10.18653/v1/W15-3049",
    pages = "392--395"
}

@inproceedings{feng-etal-2022-language,
    title = "Language-agnostic {BERT} Sentence Embedding",
    author = "Feng, Fangxiaoyu  and
      Yang, Yinfei  and
      Cer, Daniel  and
      Arivazhagan, Naveen  and
      Wang, Wei",
    editor = "Muresan, Smaranda  and
      Nakov, Preslav  and
      Villavicencio, Aline",
    booktitle = "Proceedings of the 60th Annual Meeting of the Association for Computational Linguistics (Volume 1: Long Papers)",
    month = may,
    year = "2022",
    address = "Dublin, Ireland",
    publisher = "Association for Computational Linguistics",
    url = "https://aclanthology.org/2022.acl-long.62/",
    doi = "10.18653/v1/2022.acl-long.62",
    pages = "878--891"
}

@inproceedings{finkelstein-etal-2024-introducing,
    title = "Introducing the {N}ews{P}a{LM} {MBR} and {QE} Dataset: {LLM}-Generated High-Quality Parallel Data Outperforms Traditional Web-Crawled Data",
    author = "Finkelstein, Mara  and
      Vilar, David  and
      Freitag, Markus",
    editor = "Haddow, Barry  and
      Kocmi, Tom  and
      Koehn, Philipp  and
      Monz, Christof",
    booktitle = "Proceedings of the Ninth Conference on Machine Translation",
    month = nov,
    year = "2024",
    address = "Miami, Florida, USA",
    publisher = "Association for Computational Linguistics",
    url = "https://aclanthology.org/2024.wmt-1.126/",
    doi = "10.18653/v1/2024.wmt-1.126",
    pages = "1355--1372"
}

@inproceedings{qi-etal-2020-stanza,
    title = "{S}tanza: A Python Natural Language Processing Toolkit for Many Human Languages",
    author = "Qi, Peng  and
      Zhang, Yuhao  and
      Zhang, Yuhui  and
      Bolton, Jason  and
      Manning, Christopher D.",
    editor = "Celikyilmaz, Asli  and
      Wen, Tsung-Hsien",
    booktitle = "Proceedings of the 58th Annual Meeting of the Association for Computational Linguistics: System Demonstrations",
    month = jul,
    year = "2020",
    address = "Online",
    publisher = "Association for Computational Linguistics",
    url = "https://aclanthology.org/2020.acl-demos.14/",
    doi = "10.18653/v1/2020.acl-demos.14",
    pages = "101--108"
}

@inproceedings{tiedemann-2012-parallel,
    title = "Parallel Data, Tools and Interfaces in {OPUS}",
    author = {Tiedemann, J{\"o}rg},
    editor = "Calzolari, Nicoletta  and
      Choukri, Khalid  and
      Declerck, Thierry  and
      Do{\u{g}}an, Mehmet U{\u{g}}ur  and
      Maegaard, Bente  and
      Mariani, Joseph  and
      Moreno, Asuncion  and
      Odijk, Jan  and
      Piperidis, Stelios",
    booktitle = "Proceedings of the Eighth International Conference on Language Resources and Evaluation ({LREC}'12)",
    month = may,
    year = "2012",
    address = "Istanbul, Turkey",
    publisher = "European Language Resources Association (ELRA)",
    url = "https://aclanthology.org/L12-1246/",
    pages = "2214--2218"
}

@misc{gemmateam2026gemma4,
      title={Gemma 4 Technical Report}, 
      author={Gemma Team},
      year={2026},
      eprint={2607.02770},
      archivePrefix={arXiv},
      primaryClass={cs.CL},
      url={https://arxiv.org/abs/2607.02770}, 
}

@inproceedings{kocmi-etal-2025-command,
    title = "Command-A-Translate: Raising the Bar of Machine Translation with Difficulty Filtering",
    author = {Kocmi, Tom  and
      Arkhangorodsky, Arkady  and
      Berard, Alexandre  and
      Blunsom, Phil  and
      Cahyawijaya, Samuel  and
      Dehaze, Th{\'e}o  and
      Fadaee, Marzieh  and
      Frosst, Nicholas  and
      Galle, Matthias  and
      Gomez, Aidan  and
      Govindarajan, Nithya  and
      Ko, Wei-Yin  and
      Kreutzer, Julia  and
      Marchisio, Kelly  and
      {\"U}st{\"u}n, Ahmet  and
      Vincent, Sebastian  and
      Zhang, Ivan},
    editor = "Haddow, Barry  and
      Kocmi, Tom  and
      Koehn, Philipp  and
      Monz, Christof",
    booktitle = "Proceedings of the Tenth Conference on Machine Translation",
    month = nov,
    year = "2025",
    address = "Suzhou, China",
    publisher = "Association for Computational Linguistics",
    url = "https://aclanthology.org/2025.wmt-1.55/",
    doi = "10.18653/v1/2025.wmt-1.55",
    pages = "789--799",
    ISBN = "979-8-89176-341-8"
}

@inproceedings{jaswal-2025-takes,
    title = "It Takes Two: A Dual Stage Approach for Terminology-Aware Translation",
    author = "Jaswal, Akshat",
    editor = "Haddow, Barry  and
      Kocmi, Tom  and
      Koehn, Philipp  and
      Monz, Christof",
    booktitle = "Proceedings of the Tenth Conference on Machine Translation",
    month = nov,
    year = "2025",
    address = "Suzhou, China",
    publisher = "Association for Computational Linguistics",
    url = "https://aclanthology.org/2025.wmt-1.112/",
    doi = "10.18653/v1/2025.wmt-1.112",
    pages = "1344--1350",
    ISBN = "979-8-89176-341-8"
}

@inproceedings{garcia-gilabert-etal-2025-terminology,
    title = "Terminology-Constrained Translation from Monolingual Data Using {GRPO}",
    author = "Garcia Gilabert, Javier  and
      Escolano, Carlos  and
      Liao, Xixian  and
      Melero, Maite",
    editor = "Haddow, Barry  and
      Kocmi, Tom  and
      Koehn, Philipp  and
      Monz, Christof",
    booktitle = "Proceedings of the Tenth Conference on Machine Translation",
    month = nov,
    year = "2025",
    address = "Suzhou, China",
    publisher = "Association for Computational Linguistics",
    url = "https://aclanthology.org/2025.wmt-1.111/",
    doi = "10.18653/v1/2025.wmt-1.111",
    pages = "1335--1343",
    ISBN = "979-8-89176-341-8"
}

@inproceedings{semenov-etal-2025-findings,
    title = "Findings of the {WMT}25 Terminology Translation Task: Terminology is Useful Especially for Good {MT}s",
    author = "Semenov, Kirill  and
      Huang, Xu  and
      Zouhar, Vil{\'e}m  and
      Berger, Nathaniel  and
      Zhu, Dawei  and
      Oncevay, Arturo  and
      Chen, Pinzhen",
    editor = "Haddow, Barry  and
      Kocmi, Tom  and
      Koehn, Philipp  and
      Monz, Christof",
    booktitle = "Proceedings of the Tenth Conference on Machine Translation",
    month = nov,
    year = "2025",
    address = "Suzhou, China",
    publisher = "Association for Computational Linguistics",
    url = "https://aclanthology.org/2025.wmt-1.30/",
    doi = "10.18653/v1/2025.wmt-1.30",
    pages = "554--576",
    ISBN = "979-8-89176-341-8"
}

@inproceedings{semenov-etal-2023-findings,
    title = "Findings of the {WMT} 2023 Shared Task on Machine Translation with Terminologies",
    author = "Semenov, Kirill  and
      Zouhar, Vil{\'e}m  and
      Kocmi, Tom  and
      Zhang, Dongdong  and
      Zhou, Wangchunshu  and
      Jiang, Yuchen Eleanor",
    editor = "Koehn, Philipp  and
      Haddow, Barry  and
      Kocmi, Tom  and
      Monz, Christof",
    booktitle = "Proceedings of the Eighth Conference on Machine Translation",
    month = dec,
    year = "2023",
    address = "Singapore",
    publisher = "Association for Computational Linguistics",
    url = "https://aclanthology.org/2023.wmt-1.54/",
    doi = "10.18653/v1/2023.wmt-1.54",
    pages = "663--671"
}

@inproceedings{alam-etal-2021-findings,
    title = "Findings of the {WMT} Shared Task on Machine Translation Using Terminologies",
    author = "Alam, Md Mahfuz Ibn  and
      Kvapil{\'i}kov{\'a}, Ivana  and
      Anastasopoulos, Antonios  and
      Besacier, Laurent  and
      Dinu, Georgiana  and
      Federico, Marcello  and
      Gall{\'e}, Matthias  and
      Jung, Kweonwoo  and
      Koehn, Philipp  and
      Nikoulina, Vassilina",
    editor = "Barrault, Loic  and
      Bojar, Ondrej  and
      Bougares, Fethi  and
      Chatterjee, Rajen  and
      Costa-jussa, Marta R.  and
      Federmann, Christian  and
      Fishel, Mark  and
      Fraser, Alexander  and
      Freitag, Markus  and
      Graham, Yvette  and
      Grundkiewicz, Roman  and
      Guzman, Paco  and
      Haddow, Barry  and
      Huck, Matthias  and
      Yepes, Antonio Jimeno  and
      Koehn, Philipp  and
      Kocmi, Tom  and
      Martins, Andre  and
      Morishita, Makoto  and
      Monz, Christof",
    booktitle = "Proceedings of the Sixth Conference on Machine Translation",
    month = nov,
    year = "2021",
    address = "Online",
    publisher = "Association for Computational Linguistics",
    url = "https://aclanthology.org/2021.wmt-1.69/",
    pages = "652--663"
}

@inproceedings{dinu-etal-2019-training,
    title = "Training Neural Machine Translation to Apply Terminology Constraints",
    author = "Dinu, Georgiana  and
      Mathur, Prashant  and
      Federico, Marcello  and
      Al-Onaizan, Yaser",
    editor = "Korhonen, Anna  and
      Traum, David  and
      M{\`a}rquez, Llu{\'i}s",
    booktitle = "Proceedings of the 57th Annual Meeting of the Association for Computational Linguistics",
    month = jul,
    year = "2019",
    address = "Florence, Italy",
    publisher = "Association for Computational Linguistics",
    url = "https://aclanthology.org/P19-1294/",
    doi = "10.18653/v1/P19-1294",
    pages = "3063--3068"
}

@inproceedings{hokamp-liu-2017-lexically,
    title = "Lexically Constrained Decoding for Sequence Generation Using Grid Beam Search",
    author = "Hokamp, Chris  and
      Liu, Qun",
    editor = "Barzilay, Regina  and
      Kan, Min-Yen",
    booktitle = "Proceedings of the 55th Annual Meeting of the Association for Computational Linguistics (Volume 1: Long Papers)",
    month = jul,
    year = "2017",
    address = "Vancouver, Canada",
    publisher = "Association for Computational Linguistics",
    url = "https://aclanthology.org/P17-1141/",
    doi = "10.18653/v1/P17-1141",
    pages = "1535--1546"
}

@misc{liu2024makesgooddataalignment,
      title={What Makes Good Data for Alignment? A Comprehensive Study of Automatic Data Selection in Instruction Tuning}, 
      author={Wei Liu and Weihao Zeng and Keqing He and Yong Jiang and Junxian He},
      year={2024},
      eprint={2312.15685},
      archivePrefix={arXiv},
      primaryClass={cs.CL},
      url={https://arxiv.org/abs/2312.15685}, 
}

@misc{chen2024alpagasustrainingbetteralpaca,
      title={AlpaGasus: Training A Better Alpaca with Fewer Data}, 
      author={Lichang Chen and Shiyang Li and Jun Yan and Hai Wang and Kalpa Gunaratna and Vikas Yadav and Zheng Tang and Vijay Srinivasan and Tianyi Zhou and Heng Huang and Hongxia Jin},
      year={2024},
      eprint={2307.08701},
      archivePrefix={arXiv},
      primaryClass={cs.CL},
      url={https://arxiv.org/abs/2307.08701}, 
}

@misc{zhou2023limaalignment,
      title={LIMA: Less Is More for Alignment}, 
      author={Chunting Zhou and Pengfei Liu and Puxin Xu and Srini Iyer and Jiao Sun and Yuning Mao and Xuezhe Ma and Avia Efrat and Ping Yu and Lili Yu and Susan Zhang and Gargi Ghosh and Mike Lewis and Luke Zettlemoyer and Omer Levy},
      year={2023},
      eprint={2305.11206},
      archivePrefix={arXiv},
      primaryClass={cs.CL},
      url={https://arxiv.org/abs/2305.11206}, 
}

@inproceedings{kocmi-bojar-2017-curriculum,
    title = "Curriculum Learning and Minibatch Bucketing in Neural Machine Translation",
    author = "Kocmi, Tom  and
      Bojar, Ond{\v{r}}ej",
    editor = "Mitkov, Ruslan  and
      Angelova, Galia",
    booktitle = "Proceedings of the International Conference Recent Advances in Natural Language Processing, {RANLP} 2017",
    month = sep,
    year = "2017",
    address = "Varna, Bulgaria",
    publisher = "INCOMA Ltd.",
    url = "https://aclanthology.org/R17-1050/",
    doi = "10.26615/978-954-452-049-6_050",
    pages = "379--386"
}

@inproceedings{platanios-etal-2019-competence,
    title = "Competence-based Curriculum Learning for Neural Machine Translation",
    author = "Platanios, Emmanouil Antonios  and
      Stretcu, Otilia  and
      Neubig, Graham  and
      Poczos, Barnabas  and
      Mitchell, Tom",
    editor = "Burstein, Jill  and
      Doran, Christy  and
      Solorio, Thamar",
    booktitle = "Proceedings of the 2019 Conference of the North {A}merican Chapter of the Association for Computational Linguistics: Human Language Technologies, Volume 1 (Long and Short Papers)",
    month = jun,
    year = "2019",
    address = "Minneapolis, Minnesota",
    publisher = "Association for Computational Linguistics",
    url = "https://aclanthology.org/N19-1119/",
    doi = "10.18653/v1/N19-1119",
    pages = "1162--1172"
}

@inproceedings{bengio-2009,
author = {Bengio, Yoshua and Louradour, J\'{e}r\^{o}me and Collobert, Ronan and Weston, Jason},
title = {Curriculum learning},
year = {2009},
isbn = {9781605585161},
publisher = {Association for Computing Machinery},
address = {New York, NY, USA},
url = {https://doi.org/10.1145/1553374.1553380},
doi = {10.1145/1553374.1553380},
booktitle = {Proceedings of the 26th Annual International Conference on Machine Learning},
pages = {41–48},
numpages = {8},
location = {Montreal, Quebec, Canada},
series = {ICML '09}
}

@inproceedings{van-der-wees-etal-2017-dynamic,
    title = "Dynamic Data Selection for Neural Machine Translation",
    author = "van der Wees, Marlies  and
      Bisazza, Arianna  and
      Monz, Christof",
    editor = "Palmer, Martha  and
      Hwa, Rebecca  and
      Riedel, Sebastian",
    booktitle = "Proceedings of the 2017 Conference on Empirical Methods in Natural Language Processing",
    month = sep,
    year = "2017",
    address = "Copenhagen, Denmark",
    publisher = "Association for Computational Linguistics",
    url = "https://aclanthology.org/D17-1147/",
    doi = "10.18653/v1/D17-1147",
    pages = "1400--1410"
}

@inproceedings{axelrod-etal-2011-domain,
    title = "Domain Adaptation via Pseudo In-Domain Data Selection",
    author = "Axelrod, Amittai  and
      He, Xiaodong  and
      Gao, Jianfeng",
    editor = "Barzilay, Regina  and
      Johnson, Mark",
    booktitle = "Proceedings of the 2011 Conference on Empirical Methods in Natural Language Processing",
    month = jul,
    year = "2011",
    address = "Edinburgh, Scotland, UK.",
    publisher = "Association for Computational Linguistics",
    url = "https://aclanthology.org/D11-1033/",
    pages = "355--362"
}

@article{villegas2018mespen,
  title={The MeSpEN resource for English-Spanish medical machine translation and terminologies: Census of parallel corpora, glossaries and term translations},
  author={Villegas, Marta and Intxaurrondo, Ander and Gonzalez-Agirre, Aitor and Marimon, Montserrat and Krallinger, Martin},
  journal={LREC MultilingualBIO: multilingual biomedical text processing},
  year={2018}
}

\newpage
\appendix

\section{Synthetic-data pipeline details}
\label{app:synthetic-pipeline}

\begin{figure*}[t]
\centering
\resizebox{\textwidth}{!}
{
\begin{tikzpicture}[
  font=\small,
  node distance=3.5mm and 14mm,
  box/.style={draw, rounded corners=2pt, align=center, text width=54mm,
              minimum height=7.5mm, inner sep=2.5pt, fill=gray!8},
  dat/.style={box, fill=gray!25},
  sel/.style={box, fill=orange!25},
  gem/.style={box, fill=green!18},
  fil/.style={box, fill=cyan!18},
  lab/.style={font=\small\bfseries},
  arr/.style={-{Stealth[length=2mm]}, thick},
]
\node[lab] (mtitle) {MeSpEn: terminology $\rightarrow$ text};
\node[dat, below=of mtitle] (m1) {Bilingual medical glossaries};
\node[box, below=of m1] (m2) {Clean entries, split term variants};
\node[sel, below=of m2] (m3) {SalamandraTA translates each term;\\ keep only mismatches (\textbf{hard terms})};
\node[box, below=of m3] (m4) {Group related terms\\ (LaBSE semantic clusters)};
\node[gem, below=of m4] (m5) {Gemma-4-31B writes parallel text\\ (one sentence up to multi-paragraph)};
\node[fil, below=of m5] (m6) {Rule checks + LaBSE filter\\ (term \& sentence $\geq 0.80$)};
\node[dat, below=of m6] (m7) {12{,}479 instances, 79 pairs};
\draw[arr] (m1) -- (m2); \draw[arr] (m2) -- (m3); \draw[arr] (m3) -- (m4);
\draw[arr] (m4) -- (m5); \draw[arr] (m5) -- (m6); \draw[arr] (m6) -- (m7);

\node[lab, right=76mm of mtitle] (etitle) {EMEA: text $\rightarrow$ terminology $\rightarrow$ documents};
\node[dat, below=of etitle] (e1) {EMEA parallel corpus\\ (sentence-aligned)};
\node[gem, below=of e1] (e2) {Gemma-4-31B extracts aligned\\ term pairs per sentence};
\node[sel, below=of e2] (e3) {SalamandraTA translates \emph{without} glossary; keep sentences with\\ $\geq 1$ missed term (\textbf{hard sentences})};
\node[fil, below=of e3] (e4) {Quality filters + LaBSE audit\\ (term $\geq 0.70$, sentence $\geq 0.88$)};
\node[box, below=of e4] (e5) {Concatenate into sentence /\\ paragraph / document mixture};
\node[box, below=of e5] (e6) {Lemmatize glossary\\ (text left unchanged)};
\node[dat, below=of e6] (e7) {21{,}136 instances, 19 pairs};
\draw[arr] (e1) -- (e2); \draw[arr] (e2) -- (e3); \draw[arr] (e3) -- (e4);
\draw[arr] (e4) -- (e5); \draw[arr] (e5) -- (e6); \draw[arr] (e6) -- (e7);

\node[dat, below=9mm of $(m7)!0.5!(e7)$, text width=72mm, fill=violet!15]
  (final) {Instruction-tuning data:\\ \textbf{33{,}615 instances, 94 directions, 29 languages}};
\draw[arr] (m7.south) |- ($(final.north)+(0,3mm)$) -- (final.north);
\draw[arr] (e7.south) |- ($(final.north)+(0,3mm)$) -- (final.north);

\end{tikzpicture}
}
\caption{The two synthetic-data pipelines.}
\label{fig:pipeline}
\end{figure*}
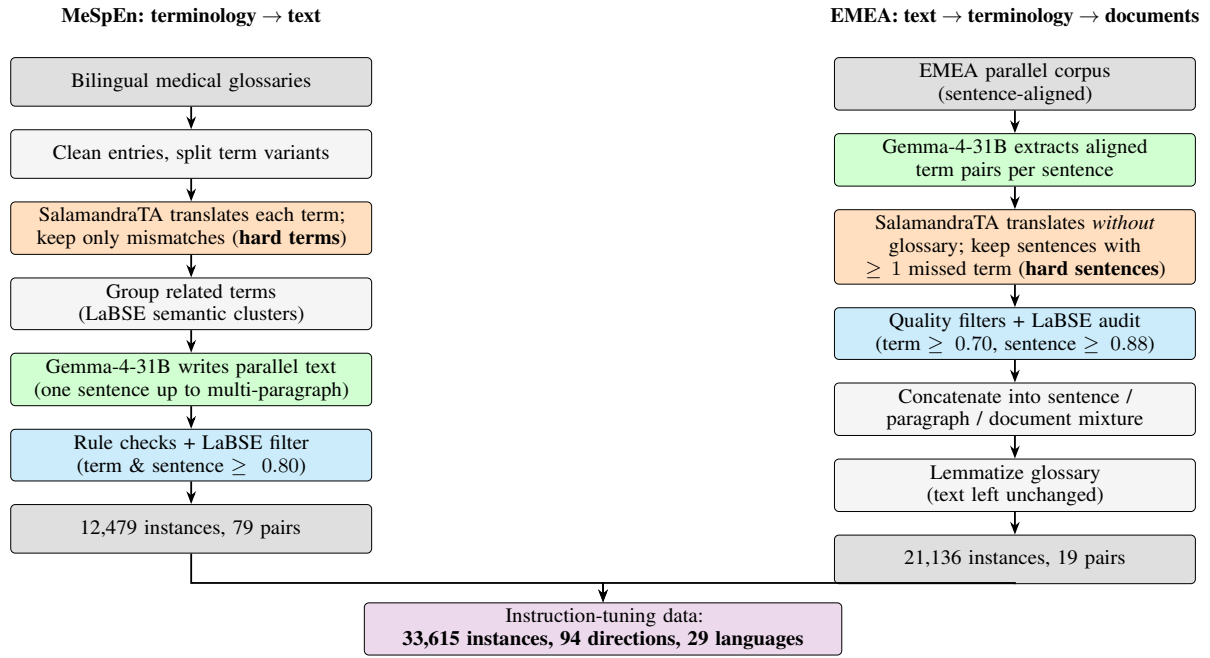

Figure~\ref{fig:pipeline} gives an overview of the two pipelines. 
MeSpEn starts from terminology and generates text around it, while EMEA pipeline starts from the EMEA corpus after cleaning with ParaCLEAN \citep{mashParaCLEANImprovingTranslation} and extracts terminology out of it.
Both contain the same two quality gates: a \emph{hard-case selection} step, where \textsc{SalamandraTA} itself decides what is worth training on, and a \emph{semantic filter} based on LaBSE similarity.


\subsection{MeSpEn: generating text from terminology}
\label{sec:mespen-data}

\paragraph{1. Glossary cleaning.}
The raw MeSpEn glossary files contain formatting noise: numbered variant lists, domain markers, phonetic transcriptions, part-of-speech tags, and occasional swapped columns. A single cleaning pass fixes these, splits multi-variant entries into pipe-separated alternatives, and deduplicates rows.

\paragraph{2. Hard-term selection.}
\textsc{SalamandraTA-7B-instruct} v2.0 translates every glossary term on its own, and the output is compared to the gold entry with a lenient matcher, so that casing, punctuation, or spacing differences do not count as errors. Terms the model already translates correctly are discarded; only genuine mismatches are used as seeds.

\paragraph{3. Semantic grouping.}
Each synthetic text is generated around a small group of glossary terms, so the terms in a group must plausibly belong to the same text: sampling them at random produces absurd combinations (e.g., an obstetrics term next to a dental one). We therefore cluster the glossary into medical sub-domains with LaBSE, and draw the terms of each group mostly (80\% of the time) from a single cluster.

\paragraph{4. Text generation.}
Gemma-4-31B receives a term group and writes the source text and its translation in one pass, at lengths ranging from a single sentence (25\%) over short passages and paragraphs (58\%) to multi-paragraph documents (17\%); short texts carry 1--3 terms, documents more. Two tricks keep the output realistic. First, the model must invent a one-sentence scenario before writing (e.g., ``a discharge summary after knee surgery''), which anchors the text in a concrete setting instead of a vague list of facts. Second, in document-length samples the main term has to appear at least twice and be translated the same way each time. The model finally reports which surface form each term actually took (say, the plural \emph{c\'elulas} rather than the glossary form \emph{c\'elula}); the next step depends on these.

\paragraph{5. Filtering.}
Two kinds of errors slip through generation. Sometimes a reported term does not actually appear in the text it claims to be in; simple rule checks catch these, and also drop degenerate samples (extreme length ratios, source copied as target, duplicates). Sometimes both source and target read fluently, yet they quietly say different things; LaBSE similarity catches these, by requiring each term pair and the full text pair to score at least $0.80$.
Surviving records are formatted as prompt--answer pairs, with document-form records oversampled ($\approx$10$\times$) to strengthen multi-paragraph layout.

\subsection{EMEA: extracting terminology from text}
\label{sec:emea-data}

\paragraph{1. Term extraction.}
Gemma-4-31B reads each aligned sentence pair and extracts the domain-specific term pairs it contains. Numeric ``terms'' (dosages such as \emph{720 IU/kg}) are removed.

\paragraph{2. Normalization.}
EMEA text carries detokenization artifacts that would otherwise end up inside glossary terms: ``\emph{Parkinson' s}'', ``\emph{13, 6 mg}'', ``\emph{mg/ ml}''. A conservative pass repairs these (``\emph{Parkinson's}'', ``\emph{13,6 mg}'', ``\emph{mg/ml}'').

\paragraph{3. Hard-sentence selection.}
\textsc{SalamandraTA} translates every source sentence \emph{without} any glossary in the prompt, and we check which gold terms appear in its output. If all of them do, the model needs no help on this sentence, so only sentences with at least one missed term are kept.

\paragraph{4. Filtering and audit.}
The same two error types as in MeSpEn are checked. Sometimes Gemma reports a term that is not literally in the sentence (e.g., one it inferred rather than read), so we look up every extracted term in its own
sentence and drop the record when it is missing, along with truncated or length-distorted samples. Sometimes the two sides quietly diverge in meaning; LaBSE similarity catches these, by requiring the sentence pair to score at least $0.88$.

\paragraph{5. Document construction.}
EMEA sentences arrive in corpus order, so consecutive sentences from the same document section can simply be concatenated back into longer units.
Each training instance is sampled from a length profile, that is, a single sentence, a short space-joined paragraph, or a document of several hundred to $\approx$1{,}000 source words with newline layout, so that the model sees every granularity.
A document's glossary is the union of its sentences' term pairs.

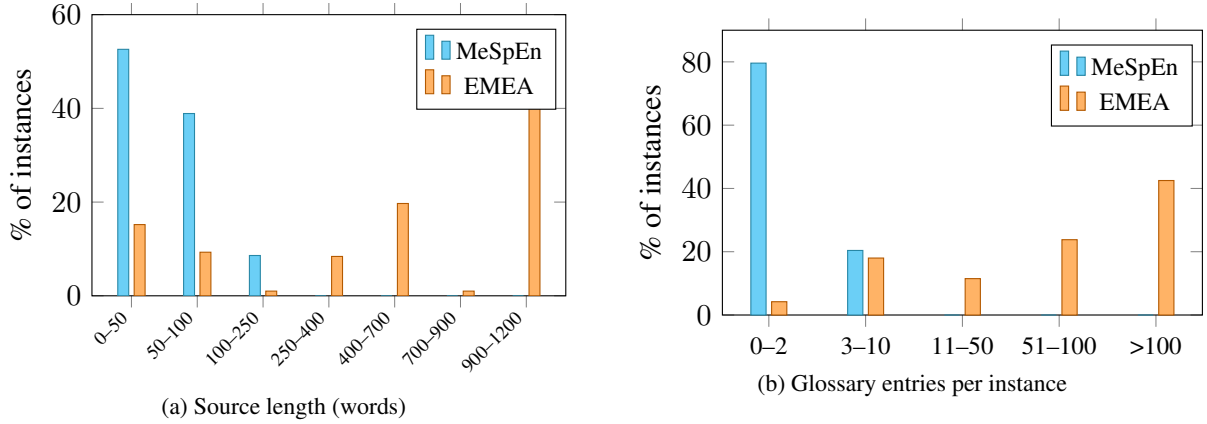
\begin{figure*}[t]
\centering
\begin{minipage}{0.48\textwidth}
\centering
\resizebox{\linewidth}{!}
{
\begin{tikzpicture}
\begin{axis}[
  ybar, bar width=4pt, width=\linewidth, height=52mm,
  symbolic x coords={0--50,50--100,100--250,250--400,400--700,700--900,900--1200},
  xtick=data, x tick label style={font=\scriptsize, rotate=45, anchor=east},
  ylabel={\% of instances}, ymin=0, ymax=60,
  legend style={font=\footnotesize, at={(0.98,0.95)}, anchor=north east},
  enlarge x limits=0.10,
]
\addplot+[fill=cyan!50, draw=cyan!60!black] coordinates
  {(0--50,52.6) (50--100,38.9) (100--250,8.6) (250--400,0) (400--700,0) (700--900,0) (900--1200,0)};
\addplot+[fill=orange!60, draw=orange!70!black] coordinates
  {(0--50,15.2) (50--100,9.3) (100--250,1.0) (250--400,8.4) (400--700,19.7) (700--900,1.0) (900--1200,45.4)};
\legend{MeSpEn, EMEA}
\end{axis}
\end{tikzpicture} }\\
\footnotesize (a) Source length (words)
\end{minipage}\hfill
\begin{minipage}{0.48\textwidth}
\centering
\resizebox{\linewidth}{!}
{
\begin{tikzpicture}
\begin{axis}[
  ybar, bar width=5.5pt, width=\linewidth, height=52mm,
  symbolic x coords={0--2,3--10,11--50,51--100,{>100}},
  xtick=data, x tick label style={font=\footnotesize},
  ylabel={\% of instances}, ymin=0, ymax=90,
  legend style={font=\footnotesize, at={(0.98,0.95)}, anchor=north east},
  enlarge x limits=0.12,
]
\addplot+[fill=cyan!50, draw=cyan!60!black]  coordinates
  {(0--2,79.6) (3--10,20.4) (11--50,0) (51--100,0) ({>100},0)};
\addplot+[fill=orange!60, draw=orange!70!black] coordinates
  {(0--2,4.2) (3--10,18.0) (11--50,11.5) (51--100,23.8) ({>100},42.5)};
\legend{MeSpEn, EMEA}
\end{axis}
\end{tikzpicture}}\\
\footnotesize (b) Glossary entries per instance
\end{minipage}
\caption{Data profile of the two synthetic datasets. }
\label{fig:dataprofile}
\end{figure*}

\paragraph{6. Glossary lemmatization.}
Finally, every glossary entry is reduced to its lemma form with Stanza \citep{qi-etal-2020-stanza} and deduplicated. The glossary thus advertises \emph{apple} $\rightarrow$ \emph{manzana} even where the text contains \emph{apples}/\emph{manzanas}, and the model must learn to inflect prescribed terms in context rather than copy surface strings. We verified that the longest formatted instances stay within the 8{,}192-token training context.

\subsection{Data profile}
\label{sec:synthetic-profile}

Figure~\ref{fig:dataprofile} summarizes the two corpora.
MeSpEn contributes short, term-dense texts in 79 language pairs (median 45 source words, usually 1--2 glossary entries), while EMEA contributes long documents in 19 pairs (median 645 source words, median 83 glossary entries).

\section{Hard/easy pool sizes per direction}
\label{app:pools}

Table~\ref{tab:pools} gives, for every direction, the size of the hard pool (v2.0 misses at least one term) and the easy pool (v2.0 already produces every term), counted after the quality and LaBSE
audit gates and before any subsampling. 
Every ratio-curve run of \S\ref{sec:ablation} draws 4{,}000 records per direction from these pools; the \emph{random} baseline draws them from the union of the two, which is why its hard share tracks the corpus's natural one.

\begin{table}[t]
\centering
\small
\setlength{\tabcolsep}{8pt}
\begin{tabular}{l rrr}
\toprule
Direction & Hard & Easy & \%Hard \\
\midrule
cs-de & 7{,}291  & 7{,}929  & 47.9 \\
cs-en & 15{,}187 & 30{,}291 & 33.4 \\
de-en & 14{,}835 & 30{,}886 & 32.4 \\
en-et & 17{,}348 & 12{,}457 & 58.2 \\
en-pl & 20{,}404 & 18{,}558 & 52.4 \\
es-bg & 32{,}219 & 32{,}839 & 49.5 \\
es-cs & 36{,}281 & 28{,}473 & 56.0 \\
es-da & 34{,}639 & 30{,}472 & 53.2 \\
es-el & 23{,}646 & 23{,}587 & 50.1 \\
es-fi & 22{,}734 & 13{,}322 & 63.1 \\
es-hu & 31{,}301 & 17{,}805 & 63.7 \\
es-lt & 24{,}345 & 19{,}242 & 55.9 \\
es-mt & 33{,}659 & 27{,}723 & 54.8 \\
es-nl & 36{,}629 & 36{,}781 & 49.9 \\
es-pl & 26{,}656 & 22{,}981 & 53.7 \\
es-ro & 42{,}265 & 29{,}028 & 59.3 \\
es-sk & 38{,}744 & 28{,}557 & 57.6 \\
es-sl & 33{,}429 & 28{,}357 & 54.1 \\
es-sv & 30{,}664 & 32{,}429 & 48.6 \\
\midrule
Total & 522{,}276 & 471{,}717 & 52.5 \\
\bottomrule
\end{tabular}
\caption{Per-direction pool sizes after quality filtering. The hard
share ranges from 32.4\% (de-en) to 63.7\% (es-hu).}
\label{tab:pools}
\end{table}

\section{Background instruction-tuning mixture}
\label{app:it-mixture}

All training runs share the same background mixture: the instruction-tuning
recipe of the public \textsc{SalamandraTA-7b-instruct} v2 release,\footnote{\url{https://huggingface.co/BSC-LT/salamandraTA-7b-instruct/tree/v2.0}} which covers general, multi-reference, paragraph-level, and document-level translation, post-editing, grammar correction, gender-fair translation, and named-entity recognition. Table~\ref{tab:it-mixture} lists every dataset with its sampling cap and instance count; per-dataset language coverage, licences, and references are given in the release model card. High-resource corpora enter the mixture through a fixed per-direction sampling cap (the ``Sample/dir.''\ column); smaller datasets are used in full. The background totals 682{,}431 instances and is identical in every run. Only the terminology-aware component added on top of it changes
(\S\ref{app:synthetic-pipeline}).

\section{Ablation: full numbers}
\label{app:ablation-numbers}

\begin{table}[h]
\centering
\small
\setlength{\tabcolsep}{4.5pt}
\begin{tabular}{l cccc}
\toprule
Mixture & en$\rightarrow$de & en$\rightarrow$es & en$\rightarrow$ru & Avg \\
\midrule
0\% hard   & 85.1 (68.1) & 80.1 (74.0) & 70.5 (57.4) & 78.7 (66.5) \\
25\%  & 90.8 (68.2) & 87.5 (74.9) & 81.3 (58.9) & 86.7 (67.4) \\
50\%   & 91.0 (68.4) & 88.1 (74.7) & 79.6 (58.0) & 86.3 (67.0) \\
random     & 93.0 (68.5) & 88.3 (74.4) & 84.1 (58.5) & 88.5 (67.1) \\
75\%   & 93.0 (68.6) & 89.9 (74.5) & 84.3 (59.0) & 89.2 (67.4) \\
100\%  & 92.3 (67.9) & 90.9 (74.6) & 86.2 (59.0) & 89.9 (67.2) \\
\bottomrule
\end{tabular}
\caption{Term accuracy (chrF in parentheses) per direction for every
point of Figure~\ref{fig:ratio-curve}. Accuracy averages are
micro-averaged over all 1{,}589 term occurrences; chrF averages are
means over the three directions.}
\end{table}

\begin{table*}[t]
\centering
\small
\begin{tabular}{llllr}
\toprule
Task & Dataset & Coverage & Sample/dir. & Instances \\
\midrule
General MT      & Tatoeba          & hundreds of directions        & 500   & 121{,}361 \\
                & NTEU             & ca/es/en $\leftrightarrow$ EU languages & 500 & 36{,}000 \\
                & EuPress          & ca/es/en $\leftrightarrow$ EU languages & 500 & 36{,}000 \\
                & WMT++            & en $\rightarrow$ 32           & 500   & 16{,}000 \\
                & FLEURS           & en/ca/es $\rightarrow$ XX     & 100   & 12{,}900 \\
Multi-reference & Tatoeba          & hundreds of directions        & 500   & 100{,}973 \\
Paragraph-level & News-Commentary  & 56 directions                 & 1{,}000 & 56{,}000 \\
                & ACAData (train)  & 96 directions                 & 500   & 27{,}852 \\
                & NewsPalm         & en $\leftrightarrow$ de       & 5{,}000 & 10{,}000 \\
                & ACAData (bench)  & 12 directions                 & full  & 5{,}944 \\
Document-level  & Europarl         & 210 directions                & 500   & 105{,}000 \\
                & Project Gutenberg (filtered) & 39 pairs          & full  & 8{,}959 \\
Post-editing    & LangMark         & en $\rightarrow$ 7            & 5{,}000 & 35{,}000 \\
                & Q21              & en $\leftrightarrow$ de/cs/lv & 5{,}000 & 30{,}000 \\
                & ApeQuest         & en $\rightarrow$ fr/nl/pt     & 5{,}000 & 15{,}000 \\
Grammar         & Synthetic noise (NTEU, EuPress) & 25 languages   & 500   & 12{,}500 \\
Gender          & EuroGEST         & en $\rightarrow$ 21           & full  & 29{,}076 \\
                & GeNTE            & en--it                        & full  & 3{,}000 \\
                & GLITTER          & en--de                        & full  & 1{,}202 \\
NER             & AnCora           & ca                            & full  & 10{,}629 \\
                & SLI NERC         & gl                            & full  & 6{,}483 \\
                & EIEC             & eu                            & full  & 2{,}552 \\
\midrule
\multicolumn{4}{l}{Total background} & 682{,}431 \\
\bottomrule
\end{tabular}
\caption{The fixed background mixture shared by all instruction-tuning runs: the recipe of the \textsc{SalamandraTA-7b-instruct} v2 release. ``Sample/dir.''\ is the per-translation-direction cap applied to high-resource corpora; ``full'' means the entire dataset is used.}
\label{tab:it-mixture}
\end{table*}

\end{document}